\documentclass{bmvc2k}
\usepackage{amsmath,amssymb,graphicx}
\usepackage{adjustbox}
\usepackage{pifont}
\usepackage{color}
\usepackage{multirow}
\usepackage{hhline}  
\usepackage{diagbox}
\usepackage{booktabs}
\usepackage{makecell}
\usepackage{colortbl}  
\usepackage{xcolor}

\newcommand{\red}[1]{\textcolor{red}{#1}}   
\newcommand{\blue}[1]{\textcolor{blue}{#1}} 

\title{Discovering an Image-Adaptive Coordinate System for Photography Processing}

\addauthor{Ziteng Cui}{cui@mi.t.u-tokyo.ac.jp}{1}
\addauthor{Lin Gu}{lin.gu@riken.jp}{2,1}
\addauthor{Tatsuya Harada}{harada@mi.t.u-tokyo.ac.jp}{1,2}

\addinstitution{
MIL Lab\\
 The University of Tokyo
}
\addinstitution{
RIKEN AIP
}

\runninghead{Ziteg Cui, Lin Gu, Tatsuya Harada}{IAC}

\begin{document}

\maketitle

\begin{abstract}
Curve $\&$ Lookup Table (LUT) based methods directly map a pixel to the target output, making them highly efficient tools for real-time photography processing. However, due to extreme memory complexity to learn full RGB space mapping, existing methods either sample a discretized 3D lattice to build a 3D LUT or decompose into three separate curves (1D LUTs) on the RGB channels. Here, we propose a novel algorithm, \textbf{IAC}, to learn an image-adaptive Cartesian coordinate system in the RGB color space before performing curve operations.
This end-to-end trainable approach enables us to efficiently adjust images with a jointly learned image-adaptive coordinate system and curves. Experimental results demonstrate that this simple strategy achieves state-of-the-art (SOTA) performance in various photography processing tasks, including photo retouching, exposure correction, and white-balance editing, while also maintaining a lightweight design and fast inference speed. 
\end{abstract}


\section{Introduction}
\label{sec:intro}

\begin{figure}[t]
    \centering
    \includegraphics[width=1.00\linewidth]{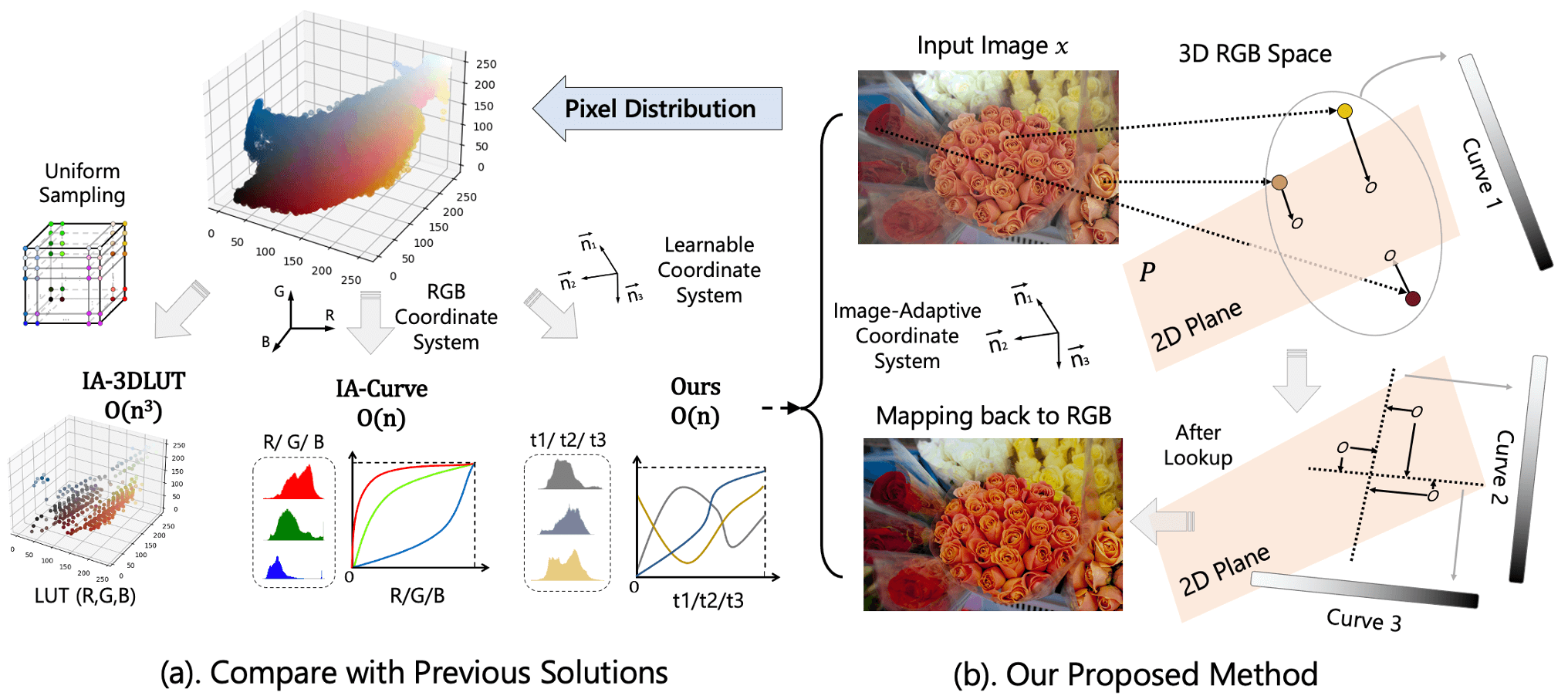}
    \caption{Compare of our \textbf{IAC} and previous image-adaptive curve $\&$ LUT methods.}
    \label{fig:comparision}
\end{figure}

The Curve $\&$ Lookup Table (LUT) serves as an array, replacing runtime computations with a simpler indexing operation. Instead of recalculating results for each operation, precomputed values stored in the table guide input values directly to corresponding outputs.
In recent years, deep network based image-adaptive curves (also referred to as 1D LUT)~\cite{moran2020curl,Curve_prediction_ECCV,song2021starenhancer,zero_dce,iccv2023_1dlut_me,vinker2021unpaired} and image-adaptive 3D LUTs~\cite{pami_3DLUT,yang2022adaint,ICME_hashlut,iccv21_3DLUT,liu20234dlut} have played a crucial role in the image processing era. Compared to methods that use networks for end-to-end mapping, curve $\&$ LUT-based methods are highly efficient and can adapt images to arbitrary scale.

Image adaptive curves (IA-curves) adjust R/ G/ B channels individually use 3 curves (3$\times$1D LUT)~\cite{Curve_prediction_ECCV,song2021starenhancer} or use a single curve to uniformly adjust all R, G, and B channels (1$\times$1D LUT)~\cite{zero_dce,iccv2023_1dlut_me}. The advantage of curve-based methods lies in their low computational cost, fast inference time, and ability to specific attribute needs (\textit{i.e.} intensity~\cite{zero_dce}). However, such operations also pose several challenges because existing curves are primarily built upon the R/ G/ B coordinate axes, which impractical to achieve adjustments for some attributes like hue and saturation~\cite{moran2020curl,iccv21_3DLUT,Heart_rate_color_space}, moreover, as it shown Fig.~\ref{fig:comparision}(a), pixel projections onto R, G, and B channels often cluster together, leading to non-uniform sampling and space wastage. 
Some efforts have been proposed to alleviate this issue, like Kim~\textit{et al.}~\cite{Curve_prediction_ECCV} add a pixel-wise local adjustment network after curve adjustment and Moran~\textit{et al.}~\cite{moran2020curl} built the curve on multi-colour space, however, these approaches typically add extra computational overhead, also building curves in multiple color spaces increases the network's learning burden.

Image-adaptive 3D LUTs (IA-3DLUTs) seem to be another solution, which quantize an RGB colour space to grid through uniform sampling (\textit{i.e.} 33*33*33, 16*16*16),
and perform lookup operations on the sampled grid. Zeng~\textit{et al.}~\cite{pami_3DLUT}
and following works~\cite{yang2022adaint,ICME_hashlut,iccv21_3DLUT}
use neural networks to learn 3D RGB cube's grid value, then use trilinear interpolation to predict missing colours in 3D RGB space. Compare to curve, 3D LUT provide more accurate colour adjustments as they operate in 3-dimensional space. However, the spatial domain sampling approach would result in many colors losing their index (ID) in 3D RGB space (see Fig.~\ref{fig:comparision}(a)). The subsequent trilinear interpolation relies on CUDA acceleration, which is often memory-intensive and not supported by commonly used frameworks like PyTorch or TFLite on mobile devices~\cite{conde2024nilut}. Meanwhile, 3D LUT methods still retain high computational complexity ($\mathcal{O}(n^3)$) and also exhibit heavier redundancy in space utilization. For instance, a 33-point image-adaptive 3D LUT typically only utilizes 5.53$\%$ of the available space.


After observing the two types of algorithms mentioned above, we thought, why not let the network learn an image-adaptive coordinate space? Where the input image is first projected into its' preference  coordinate system, and then apply  curve adjustment in the projected space before transforming back to the RGB space. In this way, we propose \textbf{IAC}, which integrate the learning of an image-adaptive coordinate space alongside curve adjustment, allows the network to adapt its coordinate space preferences for each image and task, maximizing targeted considerations while minimizing spatial complexity ($\mathcal{O}(n)$). Our solution also requires very little computational cost, and add only a small number of additional coordinate parameters compared to IA-curve methods, which easily enhance the performance and flexibility of the curve. Our contributions can be summarized as follows:

\begin{itemize}
    \item  We first applied the concept of image adaptive coordinate system, which dynamically adjust the coordinate space to better adapt to the image's own features and variations.
    
    
    \item The advantage of our method lies in its low spatial complexity ($\mathcal{O}(n)$). Meanwhile for network part, our algorithm also maintains a light-weight design ($\sim$ 39.7K parameters), making it feasible to implement on mobile and edge devices.

    \item Beyond the typical photo retouching tasks, we further validated the potential of our approach in exposure correction and white balance editing tasks. State-of-the-art (SOTA) experimental results demonstrated the effectiveness of our method.
\end{itemize}

\begin{figure}[t]
    \centering
    \includegraphics[width=1.00\linewidth]{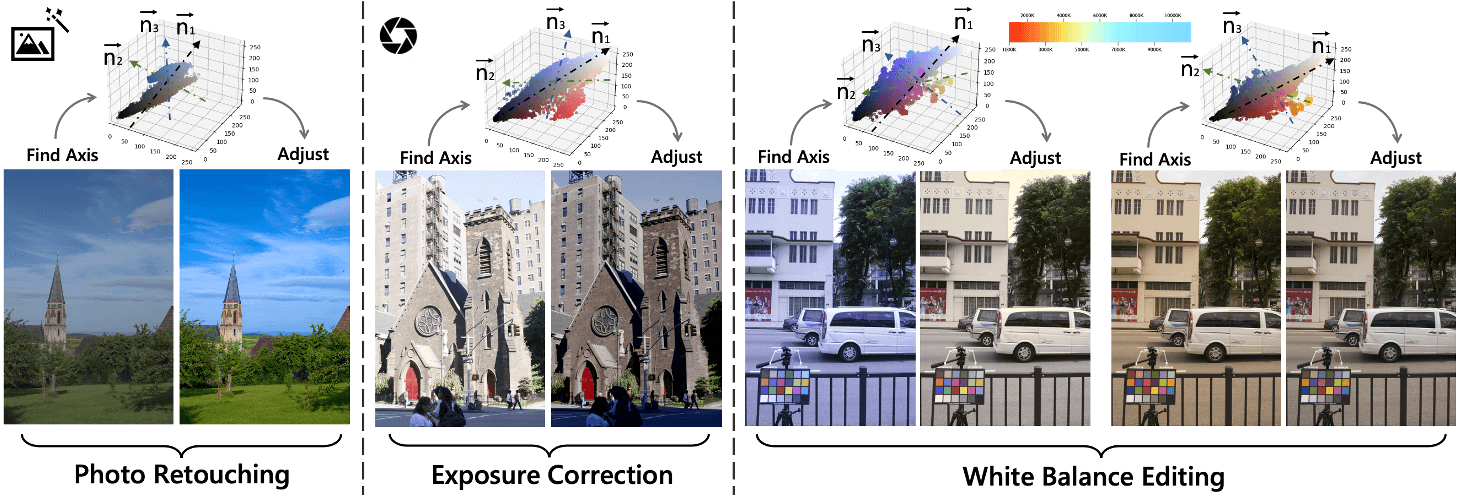}
    \caption{We adopt image-adaptive coordinate system (IAC) for various photography processing tasks, including photo retouching, exposure correction and white balance editing.}
    \label{fig:demo}

\end{figure}


\section{Photography Processing}

Photography processing aims to handle deviations occurred during the photographs capture stages and deviations introduced in the Image Signal Processor (ISP) stage. Here we primarily focus on three tasks: photo retouching, exposure correction and white balance editing, an overview of these 3 tasks is shown in Fig.~\ref{fig:demo}.

\subsection{Photo Retouching}

Photo retouching is the process of  enhancing an image to improve its appearance, clarity, or the overall quality, which is commonly used in professional photography to create visual pleasant outputs. In earlier decade, people manually adjusted photos or relied on some regression techniques~\cite{fivek_dataset} (\textit{i.e.} LASSO Regression~\cite{LASSO_regression}, Gaussian Process Regression~\cite{Gaussian_regression}). In the era of deep learning, people achieve end-to-end mapping through neural networks, employing data-driven techniques to automatically train a network capable of adjusting images~\cite{moran2020curl,Curve_prediction_ECCV,song2021starenhancer,zero_dce,RCT_ICCV21,pami_3DLUT,iccv21_3DLUT,yang2022adaint,DeepUPE_2019_CVPR,DPE_CVPR18,HDRNet,Deep_LPF,DPED,yang2023difflle,Yang_2023_ICCV,Color_representation_wacv,Cai_2023_ICCV}. Like DeepLPF~\cite{Deep_LPF} learns three different types of local parametric filters and regresses the parameters of these spatially localized filters to enhance the image, and DeepUPE~\cite{DeepUPE_2019_CVPR} introduces intermediate illumination within network to correlate the input with the anticipated enhancement results.

\subsection{Exposure Correction}

Incorrect exposure times or challenging light conditions can result images with exposure anomalies. Photo exposure correction aims handling both under $\&$ over exposure conditions  to achieve a more balanced and visually pleasing result.

Traditional exposure correction algorithms~\cite{Exposure_ECCV12,exposure_ICCV2003} prefer to use histogram adjustment to handle exposure error. When it comes to deep learning era, various neural network-based methods~\cite{Afifi_2020_CVPR_exposure_dataset,Cui_2022_BMVC,nguyen2023psenet,Ma_exposure_correction_mm23,zhou2024mslt} have been proposed to address this issue, such as Afifi~\textit{et al.}~\cite{Afifi_2020_CVPR_exposure_dataset} utilized a coarse-to-fine multi scale CNN model, and Cui~\textit{et al.}~\cite{Cui_2022_BMVC} used transformer attention to predict key ISP parameters to correct exposure. Very recently, 
Nguyen~\textit{et al.}~\cite{nguyen2023psenet} designed to use a pseudo ground-truth learning way to achieve unsupervised exposure correction. 

\subsection{White Balance Editing}

White balance (WB) editing aims to correct the acquired sRGB images with wrong  white balance setting. This task is more challenging than regular WB correction task which operated on raw-RGB, since illumination estimation is achieved on raw-RGB, once the WB setting is chosen there still remains various other non-linear operation stages in ISP, 
these operations can increase the complexity of white balance correction.~\cite{afifi2019colour_knn,Mobile_Computational}. 

Afifi~\textit{et al.}~\cite{afifi2019colour_knn} first introduced this task and proposed a method based on k-nearest neighbor (KNN) to compute a nonlinear colour mapping function for correcting images. After that, various deep learning methods~\cite{WB_CNN_CVPR2020,WB_WACV2023,WB_ICCVW2023} have been introduced in this area, these works aim to accomplish the mapping from error white-balance (WB) images to correct WB images through an end-to-end network based approach.

In this paper, we will evaluate our method on the aforementioned three tasks to confirm the effectiveness of our image-adaptive coordinate system in diverse scenarios.  Experimental results demonstrate that our approach achieves state-of-the-art (SOTA) performance across all three tasks while maintaining time and parameter efficiency.

\section{Image-Adaptive Coordinate System}
\label{methods}
An overview of \textbf{IAC} algorithm is shown in Fig~\ref{fig:comparision}(b). Given an image $x (r, g, b)$ in the RGB colour space $\mathbb{I}(R, G, B)$, our goal is to learn $x$'s most suitable coordinate projection vectors \{$\Vec{n_1}, \Vec{n_2}, \Vec{n_3}$\}, image $x$ would be mapped along \{$\Vec{n_1}, \Vec{n_2}, \Vec{n_3}$\} to projected in new coordinate system, then adjusted with 3 curves on new coordinate system. Both of the projection vector \{$\Vec{n_1}, \Vec{n_2}, \Vec{n_3}$\} and curves \{$\textit{curve}_1, \textit{curve}_2, \textit{curve}_3$\} are learned from network $\mathbb{N}$.

\subsection{Methodology}
\label{detail_methods}
\begin{figure*}[t]
    \centering
    \includegraphics[width=0.95\linewidth]{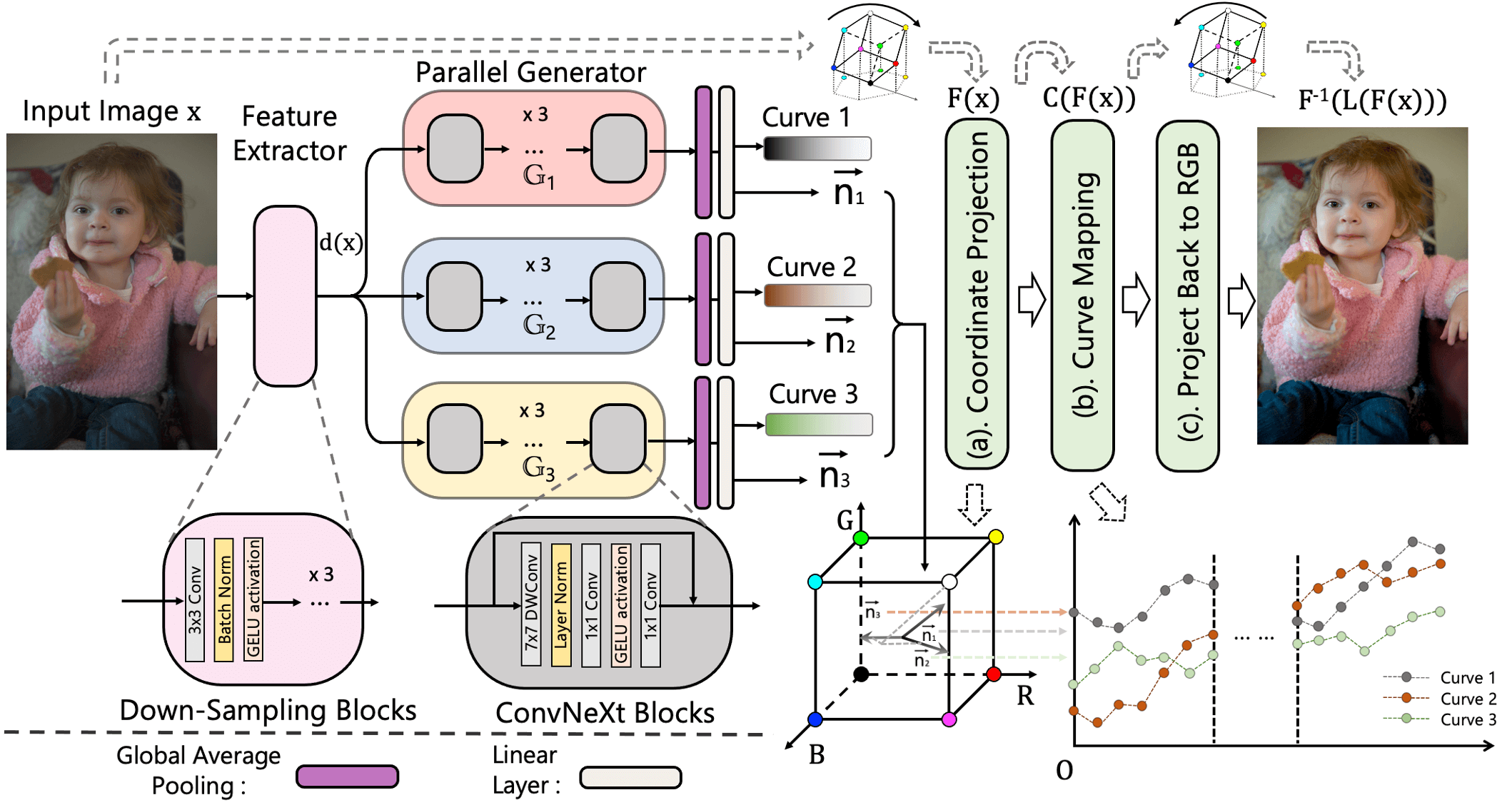}
    \caption{An overview of \textbf{IAC}'s network architecture, image $x$ would pass by network $\mathbb{N}$ to generate coordinate system vectors \{$\Vec{n_1}, \Vec{n_2}, \Vec{n_3}$\} and curves \{$\textit{curve}_1, \textit{curve}_2, \textit{curve}_3$\}.}
    \label{fig:network}
    
\end{figure*}

Image-adaptive coordinate
system's vectors $\Vec{n_1}$, $\Vec{n_2}$ and $\Vec{n_3}$ are three 3-dimensional linearly independent vectors, which  form a $3 \times 3$ invertible matrix:

\begin{equation}
 	\left[\Vec{n_1}, \Vec{n_2}, \Vec{n_3} \right] = \begin{bmatrix}
   a_1 & a_2 & a_3  \\
   b_1 & b_2 & b_3 \\
   c_1 & c_2 & c_3
  \end{bmatrix},
\end{equation}
where the RGB colour space could be seem as a special case when $\left[\Vec{n_1}, \Vec{n_2}, \Vec{n_3} \right]$ is an identity matrix. Input image $x(r, g, b) \in (H, W, 3)$ would multiply to matrix $\left[\Vec{n_1}, \Vec{n_2}, \Vec{n_3} \right]$ and project onto the new coordinates space (depicted as (a) in Fig.\ref{fig:network}):

\begin{equation}
\begin{aligned}
  F(x)  &= x \cdot \left[\Vec{n_1}, \Vec{n_2}, \Vec{n_3} \right]
  = \left[x \cdot \Vec{n_1}, x \cdot \Vec{n_2}, x \cdot \Vec{n_3} \right] \\
  &= \left[x(r), x(g), x(b) \right] \cdot \begin{bmatrix}
   a_1 & a_2 & a_3  \\
   b_1 & b_2 & b_3 \\
   c_1 & c_2 & c_3 
  \end{bmatrix}
  = \left[t_1, t_2, t_3 \right] \\
  where &: \\
  t_i &= x(r) \cdot a_i + x(g) \cdot b_i + x(b) \cdot c_i \quad i \in (1,2,3). \\
\end{aligned}
\label{eq:trans}
\end{equation}


Following Eq.\ref{eq:trans}, the image $x(r, g, b)$ would be projected with the matrix $\left[\Vec{n_1}, \Vec{n_2}, \Vec{n_3} \right]$ to $F(x)$. The projection results are represented as $F(x)(t_1, t_2, t_3)$, where $t_1, t_2, t_3$ are the projected values in the new coordinate space. Then, adaptive curves ${\textit{curve}_1, \textit{curve}_2, \textit{curve}_3}$ would be built on the $t_1, t_2, t_3$ channels to adjust the value of $F(x)$ (depicted as (b) in Fig.\ref{fig:network}). Here, we normalize the range of $t_1, t_2, t_3$ between 0 and 1. Meanwhile, the curves also range from 0 to 1, and for each curve, we designed it as 200 dimensions~\footnote{For more details, please refer to our supplementary part.}. Then we adjust the pixel value through curves \{$\textit{curve}_1, \textit{curve}_2, \textit{curve}_3$\}:

\begin{equation}
 	t_i' = \textit{curve}_i(t_i) \quad i \in (1,2,3),
\end{equation}
values in $F(x)(t_1, t_2, t_3)$ after curve mapping adjustment would be $L(F(x))(t_1', t_2', t_3')$, 
here we first perform a denormalization to recover $L(F(x))$'s normalized value, then multiply to  $\left[\Vec{n_1}, \Vec{n_2}, \Vec{n_3} \right]$'s inverse matrix to map $L(F(x))$ back to RGB colour space and get final results (depicted as (c) in Fig.\ref{fig:network}):

\begin{equation}
 	F^{-1}(L(F(x))) = L(F(x)) \cdot \left[\Vec{n_1}, \Vec{n_2}, \Vec{n_3} \right]^{-1}.
\end{equation}

We will initialize the matrix $\left[\Vec{n_1}, \Vec{n_2}, \Vec{n_3} \right]$ as an invertible matrix, and in order for the matrix $\left[\Vec{n_1}, \Vec{n_2}, \Vec{n_3} \right]$ to be invertible during learning stage, if the rank of the learned matrix $\left[\Vec{n_1}, \Vec{n_2}, \Vec{n_3} \right]$ less than 3, we will add a set of small random numbers to help it recover to rank 3. 

Afterwards, we will compute loss function (\textit{i.e.} L1 loss) between $F^{-1}(L(F(x)))$  and the ground truth $x_{gt}$ to optimize network $\mathbb{N}$, then help us find the most suitable image-adaptive coordinate projection vectors $\left[\Vec{n_1}, \Vec{n_2}, \Vec{n_3} \right]$ and curves.


\subsection{Network Design}

In this section we introduce \textbf{IAC}'s network design, network
$\mathbb{N}$ is responsible to predict image-adaptive coordinate system  \{$\Vec{n_1}, \Vec{n_2}, \Vec{n_3}$\} and curves \{$\textit{curve}_1, \textit{curve}_2, \textit{curve}_3$\}, additionally \textbf{IAC} is a general approach that could also be implemented in other framework.

As shown it in Fig.~\ref{fig:network}, input image $x$ first passes through 3 down-sampling blocks, each down-sampling block consists of a down-sampling ($\downarrow$2) 3$\times$3 convolution, batch normalization and a GELU activation~\cite{GELU}. After down-sampling process, we designed a parallel generator to predict image adaptive coordinate and curves. The parallel generator comprises three parallel branches: $\mathbb{G}_{1}$, $\mathbb{G}_{2}$, $\mathbb{G}_{3}$, each branch consists of several ConvNext~\cite{liu2022convnet} blocks, where we set the block number to 3 in our experiments, for each  ConvNext~\cite{liu2022convnet} block, it consists of a 7$\times$7 depth-wise convolution, two 1$\times$1 convolutions, layer normalization (LN), and GELU activation~\cite{GELU}, channel number of convolution blocks is set to 32, also the entire structure is linked by a residual operation. The large convolution design allows \textbf{IAC} to more effectively extract image features, which also enabling us to better acquire the global-wise image information.

Among three parallel branches: $\mathbb{G}_{1}$, $\mathbb{G}_{2}$, $\mathbb{G}_{3}$, branch $\mathbb{G}_{1}$ is responsible to predict vector $\Vec{n_1}$ and $\textit{curve}_1$, feature pass by $\mathbb{G}_1$'s ConvNext blocks would go through a global average pooling layer and linear layer to predict $\Vec{n_1} = [a_1, b_1, c_1]$ and $\textit{curve}_1$. Similarity, branch $\mathbb{G}_{2}$ is responsible to predict $\Vec{n_2}$ and $\textit{curve}_2$ and branch $\mathbb{G}_{3}$ is responsible to predict $\Vec{n_3}$ and $\textit{curve}_3$. After that the predicted vectors and curves would process input image $x$, as we mentioned in Sec.~\ref{detail_methods}. Please refer to our supplementary for more structure details.


\section{Experiments}

In this section, we selected 3 tasks to validate the effectiveness of our image-adaptive coordinate (\textbf{IAC}) method, including $(a)$. photo retouching,  $(b)$. exposure correction and  $(c)$ white balance editing. We would detailed introduce the experiments as follow~\footnote{For more experimental details and training setting, please refer to our supplementary part.}:

\begin{figure*}[]
    \centering
    \includegraphics[width=1.00\linewidth]{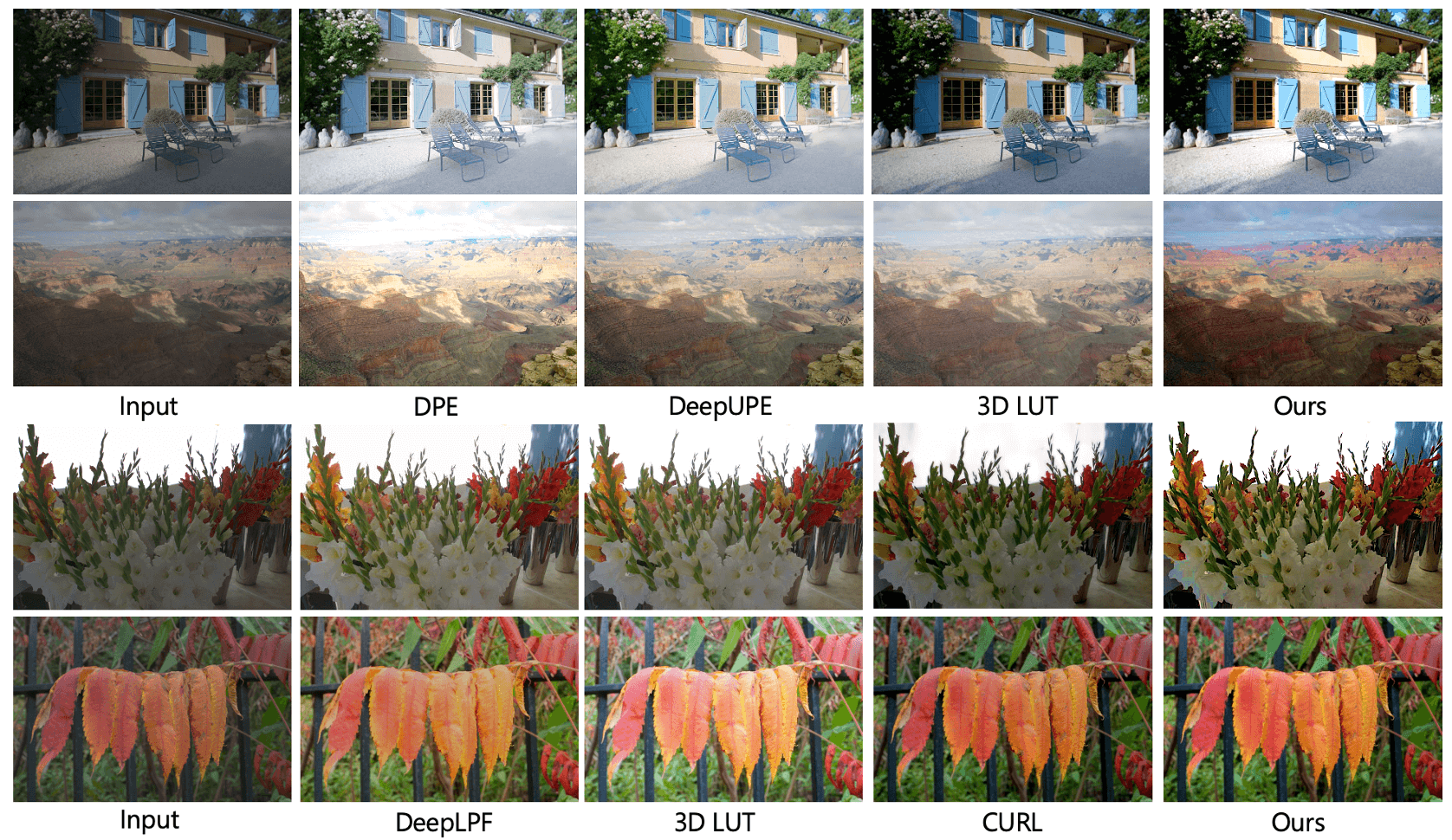}
    \caption{Visulization of photo retouching results on MIT-Adobe FiveK dataset~\cite{fivek_dataset}.}
    \label{fig:enhancement}
   
\end{figure*}

\subsection{Photo Retouching Experiments}

\begin{table}[t]
\caption{Experimental results on MIT-Adobe FiveK~\cite{fivek_dataset} dataset, we compare PSNR$\uparrow$, SSIM$\uparrow$,  parameter number ($\#$ Para)$\downarrow$ and inference time$\downarrow$. Here best results are marked as red and best results without CUDA operation are marked as blue.}
\centering
\renewcommand\arraystretch{1.4}
\begin{adjustbox}{max width = 1.0\linewidth}
\begin{tabular}{c|ccccccccc}
\hline
\hline
               & DeepUPE & DPE    & HDRNet & DeepLPF & DPED  & 3D LUT  & AdaInt & CURL   & \textbf{IAC (Ours)}  \\ \hline
PSNR           & 21.88   & 23.75  & 24.66  & 24.73   & 21.76 & 25.29  & \red{25.49}  & 24.04  & \blue{25.02}  \\ \hline
SSIM           & 0.853   & 0.828  & 0.875  & 0.916   & 0.871 & 0.922  & \red{0.926}  & 0.900  & \blue{0.902}  \\ \hline
Need CUDA ?     &   \ding{56} &  \ding{56}  &  \ding{56} & \ding{56}  & \ding{56} &  \ding{52}  &   \ding{52}     &    \ding{56}    &  \ding{56}   \\ \hline
\# Para        & 927.1K  & 3.4M   & 483.1K & 1.7M    & -     & 593.7K & 619.7K & 1.4M   & \red{39.7K}  \\ \hline
Inference Time & 0.628s  & 0.534s & 0.673s & 1.287s  & -     & \red{0.012s} & 0.018s & 0.834s & \blue{0.014s} \\ \hline
\hline
\end{tabular}
\end{adjustbox}
\label{tab:MIT5K}

\end{table}

We conclude photo retouching experiments on MIT-Adobe FiveK~\cite{fivek_dataset} dataset, MIT-Adobe FiveK dataset consists of 5000 images, each meticulously adjusted by five  experts (A/B/C/D/E). Following previous works~\cite{Deep_LPF,DeepUPE_2019_CVPR,pami_3DLUT}, we employ images adjusted by expert C as the ground truth references. And compare our methods with various SOTA photo retouching methods include DeepUPE~\cite{DeepUPE_2019_CVPR}, DPE~\cite{DPE_CVPR18}, HDRNet~\cite{HDRNet}, DeepLPF~\cite{HDRNet}, DPED~\cite{DPED}, CURL~\cite{moran2020curl}, 3D LUT~\cite{pami_3DLUT} and AdaInt~\cite{yang2022adaint}, it worth to note that 3D LUT and AdaInt are 2 image-adaptive 3D LUT methods which must rely CUDA to acceleration. 
Comparison results are shown in Table.~\ref{tab:MIT5K}, our \textbf{IAC} approach could gain best image quality performance (PSNR, SSIM) among non-CUDA methods, meanwhile keep an extremely lightweight design (39.7K parameters) and fast inference time. Some  visualization examples are shown in Fig.~\ref{fig:enhancement}, 
our \textbf{IAC} could produces ideal colour restoration results, without excessively low or brightened enhanced images, ensuring visual quality in accordance with human perception.

\begin{table*}[t]
\caption{Comparison results on the exposure correction  \textbf{ME}~\cite{Afifi_2020_CVPR_exposure_dataset} dataset, where the best results are marked as bold and second best results are marked as underline.}
\centering
\renewcommand\arraystretch{1.4}
\setlength\tabcolsep{2pt}
\begin{adjustbox}{max width = 1.0\linewidth}
\begin{tabular}{l|cc|cc|cc|cc|cc|cc|c}
\toprule
\toprule
\multirow{2}{*}{Method} & \multicolumn{2}{c|}{Expert A} & \multicolumn{2}{c|}{Expert B} & \multicolumn{2}{c|}{Expert C} & \multicolumn{2}{c|}{Expert D} & \multicolumn{2}{c|}{Expert E}      & \multicolumn{2}{c|}{Avg} & \multirow{2}{*}{Test Time $\downarrow$} \\ \cline{2-13}
& PSNR$\uparrow$ & SSIM$\uparrow$  & PSNR$\uparrow$ & SSIM$\uparrow$          & PSNR$\uparrow$ & SSIM$\uparrow$   & PSNR$\uparrow$   & SSIM$\uparrow$   & PSNR$\uparrow$  & SSIM$\uparrow$ & PSNR$\uparrow$& SSIM$\uparrow$  & \\
\Xhline{0.6pt}
HE~\cite{DIP}                      & 16.14         & 0.685         & 16.28         & 0.671         & 16.52         & 0.696         & 16.63         & 0.668         & 17.30 & 0.688 & 16.58       & 0.682      & 0.50s               \\
LIME~\cite{LIME}                    & 11.15         & 0.590          & 11.83         & 0.610         & 11.52         & 0.607         & 12.64         & 0.628         & 13.61 & 0.653 & 12.15       & 0.618      & 10.32s               \\

RetinexNet~\cite{LOL_dataset}              & 10.76         & 0.585         & 11.61         & 0.596         & 11.13         & 0.605         & 11.99         & 0.615         & 12.67 & 0.636 & 11.63       & 0.607      & 1.08s              \\
Deep-UPE~\cite{DeepUPE_2019_CVPR}                & 13.16         & 0.610         & 13.90         & 0.642         & 13.69         & 0.632         & 14.80         & 0.649         & 15.68 & 0.667 & 14.25       & 0.640      & 0.78s            \\
Zero-DCE~\cite{zero_dce}                & 11.64         & 0.536         & 12.56         & 0.539         & 12.06         & 0.544         & 12.96         & 0.548         & 13.77 & 0.580 & 12.60       & 0.549      & \textbf{0.04s}               \\
3D-LUT~\cite{pami_3DLUT}                & 13.68         & 0.591         & 11.86         & 0.577         & 12.79         & 0.627         & 12.96         & 0.548         & 14.51 & 0.602 & 13.06       & 0.519      & 0.28s               \\
SCI~\cite{ma2022sci}                    & 16.11          & 0.737         &  17.15         & 0.805         & 16.36         & 0.764         & 16.51          & 0.766        & 16.09  & 0.761 & 16.44  & 0.767      & 0.17s             \\
MSEC~\cite{Afifi_2020_CVPR_exposure_dataset}                    & 19.16         & 0.746         & 20.10         & 0.734         & 20.20         & 0.769         & 18.98         & 0.719         & 18.98 & 0.727 & 19.48       & 0.739      & 0.72s               \\
IAT~\cite{Cui_2022_BMVC}                    & \underline{19.63}        & \underline{0.780}         & 21.21        & \underline{0.816}         & 21.21         & \underline{0.820}  & 19.58   & \underline{0.805}    & 19.21 & \underline{0.797} &  20.07  & \underline{0.804}      & 0.11s              \\ 

PSENet~\cite{nguyen2023psenet}                    & 19.90        & 0.817         & 21.65        & 0.867         & 21.23         & 0.850  & 19.86   & 0.844    & 19.34 & 0.840 & 20.34       & 0.844      & 0.28s              \\ 

MSLT~\cite{zhou2024mslt}                    & 20.21        & 0.805         & \textbf{22.47}        & 0.864        & \underline{22.03}         & 0.844  & \textbf{20.33}   & 0.830    & \underline{20.04} & 0.832 & \underline{21.02}  & 0.835     & 0.24s               \\ 

\Xhline{0.6pt}

\textbf{IAC (Ours)}                     & \textbf{21.23}        & \textbf{0.829}         & \underline{21.84}        & \textbf{0.870}         & \textbf{22.05}         & \textbf{0.859}  & \underline{20.09}   & \textbf{0.846}    & \textbf{20.88} & \textbf{0.848} & \textbf{21.22}       & \textbf{0.850}      & \underline{0.09s}               \\ 
\bottomrule
\bottomrule
\end{tabular}
\end{adjustbox}
\label{tab:Exposure}

\end{table*}

\subsection{Exposure Correction Experiments}

Secondly, we conducted experiments in exposure correction task to further verify our method's effectiveness. Here we choose \textbf{ME} dataset~\cite{Afifi_2020_CVPR_exposure_dataset} which contains 24,330 8-bit sRGB images, and divided into 17,675 training images, 750 validation images, and 5905 test images. \textbf{ME} dataset is rendered from MIT-Adobe FiveK~\cite{fivek_dataset} dataset's RAW data with 5 different  exposure values (EVs), where EVs ranging from \{-1.5, -1, 0, +1, +1.5\}, including under-exposure to over-exposure conditions. This task aims to assess the model's capability to simultaneously adjust both under $\&$ over-exposure conditions.

\begin{figure*}[]
    \centering
    \includegraphics[width=1.00\linewidth]{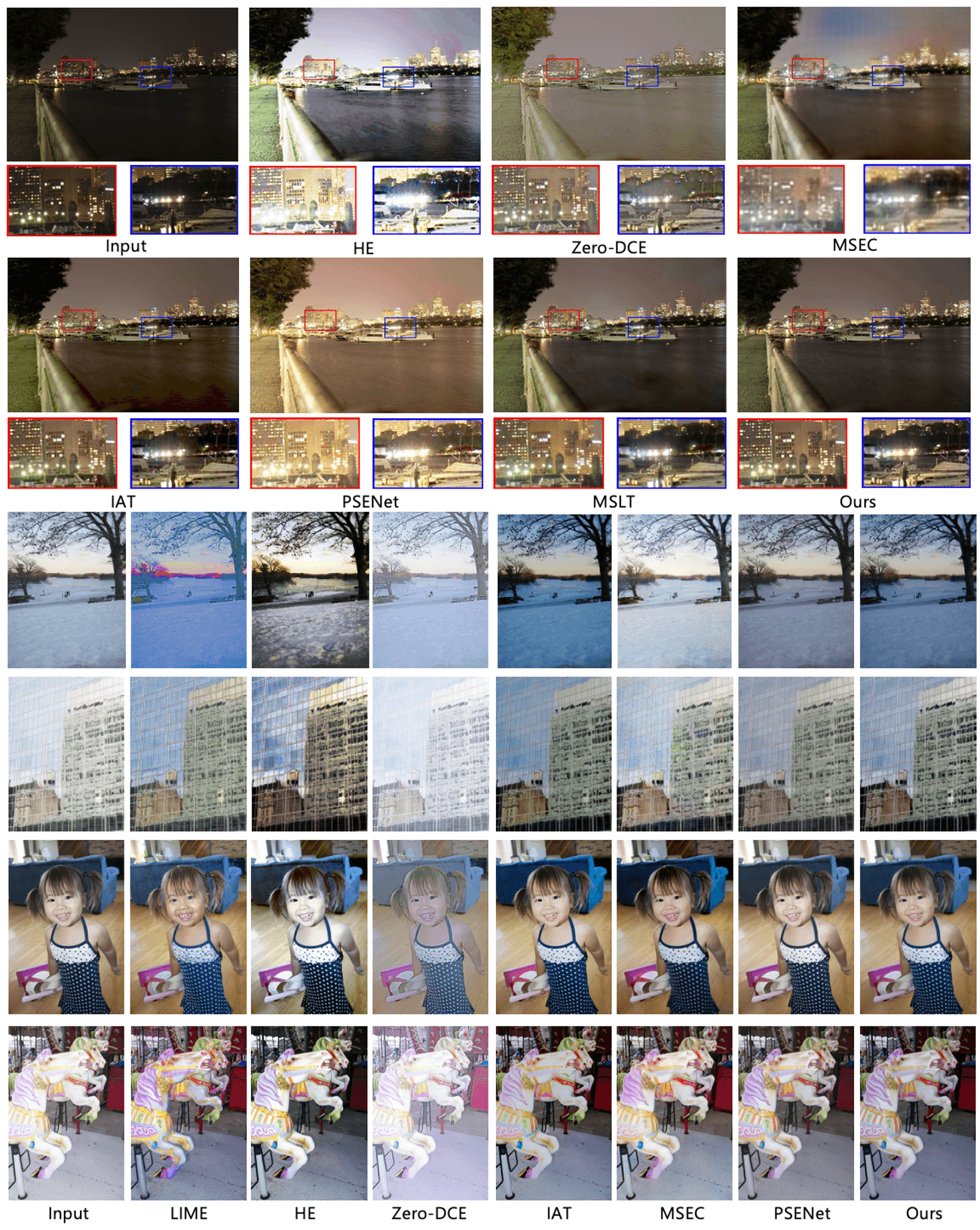}
    \caption{Visualization results on exposure correction dataset~\cite{Afifi_2020_CVPR_exposure_dataset}, line 1 and line 2 are the under-exposure correction results, meanwhile line 3 to line 6 are the over-exposure correction results, our method could both handle over$\&$under-exposure meanwhile keep more details.}
    \label{fig:exposure}
\end{figure*}

We show the experimental results in Table.~\ref{tab:Exposure}, we compare \textbf{IAC} with various methods, including traditional methods histogram equalization (HE)~\cite{DIP} and LIME~\cite{LIME}, SOTA deep-network based image enhancement methods~\cite{LOL_dataset,DeepUPE_2019_CVPR,zero_dce,ma2022sci} and SOTA deep-network based exposure correction methods~\cite{Afifi_2020_CVPR_exposure_dataset,Cui_2022_BMVC,nguyen2023psenet,zhou2024mslt}. From Table.~\ref{tab:Exposure} we can see that our method gain best performance in overall PSNR and SSIM, meanwhile keep a fast inference speed.
We also show the visulization results in Fig.~\ref{fig:exposure}, our \textbf{IAC} demonstrates the capability to effectively correct overexposure and enhance underexposure, while also efficiently preserving image details. An example in underexposure "Night" scene (Fig.~\ref{fig:exposure} line 1$\sim$2) shows that, Zero-DCE~\cite{zero_dce}, MSEC~\cite{Afifi_2020_CVPR_exposure_dataset}, and PSENet~\cite{nguyen2023psenet} tend to over-brighten images, potentially causing them to lose details, meanwhile IAT~\cite{Cui_2022_BMVC} and MSLT~\cite{zhou2024mslt} may result in low clarity.

\subsection{White Balance Editing Experiments}

For white balance editing task, we utilize the Rendered WB dataset created by Afifi~\textit{et al.}~\cite{afifi2019colour_knn}, this dataset comprises two subsets: \textbf{Set1}, containing 62,535 images captured by seven distinct DSLR cameras, and \textbf{Set2}, containing 2,881 images captured by one DSLR camera and four different phone cameras. Here we follow previous works'~\cite{afifi2019colour_knn,WB_CNN_CVPR2020} setting, which take \textbf{Set1} for training and use \textbf{Set2} for testing, same as previous work~\cite{WB_CNN_CVPR2020}, we randomly choose 12,000 images in \textbf{Set1} as the training set. 

\begin{figure*}[]
    \centering
    \includegraphics[width=1.00\linewidth]{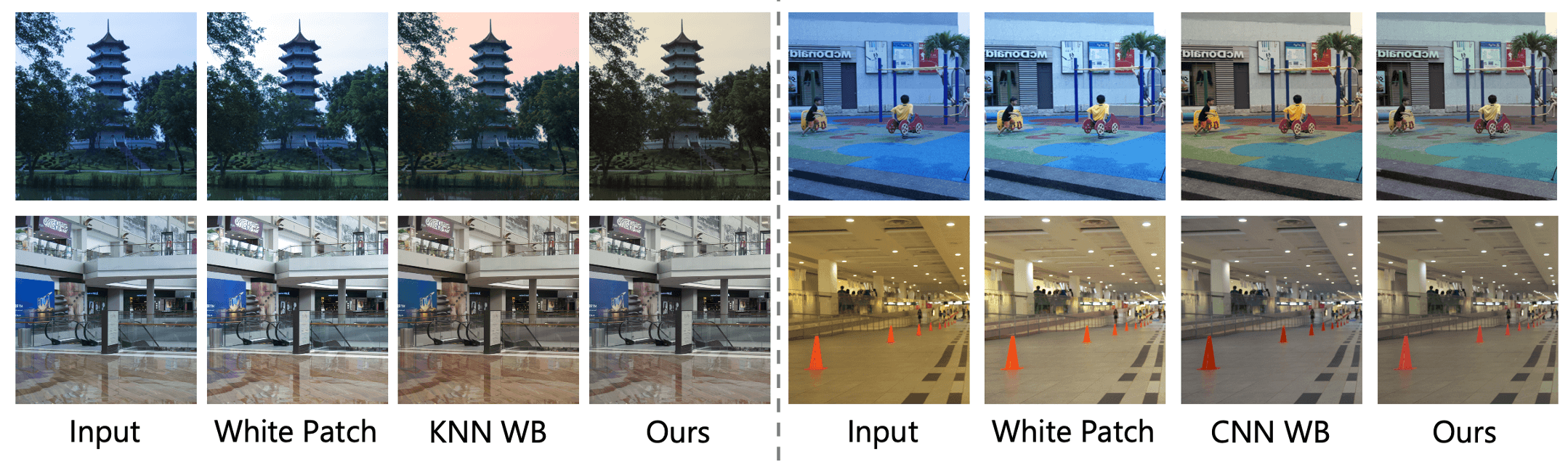}
    
    \caption{Visulization of white balance editing results on dataset~\cite{afifi2019colour_knn}'s Set2.}
    \label{fig:WB}
    
\end{figure*}

We made comparison with various WB editing methods, including the classical White Patch~\cite{whitepatch} method, along with recent methods FC4~\cite{hu2017fc4}, KNN-WB~\cite{afifi2019colour_knn} and CNN-WB~\cite{WB_CNN_CVPR2020}. Comparison results are shown in Table.~\ref{tab:wb}, 
we can see \textbf{IAC} can achieve competitive results meanwhile keep fastest inference speed. Additionally our method is much more light-weight than CNN-WB~\cite{WB_CNN_CVPR2020} (\textbf{IAC} $\sim$ 39.7K \textit{v.s.} CNN-WB $\sim$ 10M), some visualization results in Set2 are shown in Fig.~\ref{fig:WB}, demonstrates that \textbf{IAC} could also handle white balance editing task.

\begin{table*}[t]
\caption{Comparison results on the white balance editing dataset~\cite{afifi2019colour_knn}'s \textbf{Set2}, yellow colour shows best results and blue colour shows second best results.}
\centering
\renewcommand\arraystretch{1.3}
\begin{adjustbox}{max width = 1.0\linewidth}
\begin{tabular}{l|cccc|cccc|cccc|c}
\toprule
\toprule
\multirow{2}{*}{Methods} & \multicolumn{4}{c|}{MSE  $\downarrow$}          & \multicolumn{4}{c|}{MAE  $\downarrow$}     & \multicolumn{4}{c|}{Delta E 2000  $\downarrow$} & \multirow{2}{*}{Inference Time} \\ \cline{2-13}
                         & Mean   & Q1     & Q2     & Q3     & Mean  & Q1   & Q2    & Q3    & Mean    & Q1    & Q2     & Q3     &                                 \\ \hline
White Patch              & 586.72 & 148.65 & 335.76 & 664.41 & 11.26 & 6.28 & 10.17 & 16.89 & 12.28   & 8.79  & 12.07  & 15.01  & \cellcolor[HTML]{ECF4FF} 0.15s                           \\ \hline
FC4                      & 505.30 & 142.46 & 307.77 & 635.35 & 10.37 & 5.94 & 9.42  & 14.04 & 10.82   & 7.39  & 10.64  & 13.77  & 0.89s                           \\ \hline
KNN-WB                   & 171.09 & 37.04  & 87.04  & 190.88 & 4.48  & 2.26 & 3.64  & 5.95  & 5.60    & 3.43  & 4.90   & 7.06   & 0.54s                           \\ \hline
CNN-WB                   & \cellcolor{yellow!10}124.97 & \cellcolor{yellow!10}30.13  & \cellcolor{yellow!10}76.32  & \cellcolor{yellow!10}154.44 & \cellcolor{yellow!10}3.75  & \cellcolor[HTML]{ECF4FF} 2.02 & \cellcolor{yellow!10}3.08  & \cellcolor{yellow!10}4.72  & \cellcolor{yellow!10}4.90    & \cellcolor{yellow!10}3.13  & \cellcolor{yellow!10}4.35   & \cellcolor{yellow!10}6.08   & 1.2s                            \\ \hline
Ours                     & \cellcolor[HTML]{ECF4FF}130.58 & \cellcolor[HTML]{ECF4FF}33.22  & \cellcolor[HTML]{ECF4FF}72.56  & \cellcolor[HTML]{ECF4FF}180.48 & \cellcolor[HTML]{ECF4FF}3.99  & \cellcolor{yellow!10}1.98 & \cellcolor[HTML]{ECF4FF}3.44  & \cellcolor[HTML]{ECF4FF}4.87  & \cellcolor[HTML]{ECF4FF}5.13    & \cellcolor[HTML]{ECF4FF}3.24  & \cellcolor[HTML]{ECF4FF}4.48   & \cellcolor[HTML]{ECF4FF}6.27   & \cellcolor{yellow!10}0.05s                           \\ \bottomrule \bottomrule
\end{tabular}
\end{adjustbox}
\label{tab:wb}

\end{table*}

\section{Conclusion}

We present \textbf{IAC}, which learns an image-adaptive coordinate system for various photography processing tasks. Experimental results on photo retouching, exposure correction, and white balance editing showcase the superior performance of our method. In the future, we aim to extend the coordinate transformation solution to curved surface, our algorithm may yield even better results in non-uniform coordinate spaces, we also want validate the effectiveness of \textbf{IAC} for 3D application, such as 3D reconstruction in challenging lighting conditions~\cite{cui_aleth_nerf}.

\section{Acknowledgement}

This work was partially supported by JST Moonshot R$\&$D Grant Number JPMJPS2011, CREST Grant Number JPMJCR2015 and Basic Research Grant (Super AI) of Institute for AI and Beyond of the University of Tokyo.

\bibliography{egbib}

\clearpage
\appendix

\renewcommand\thesection{\Alph{section}}
\renewcommand{\thetable}{\Roman{table}} 
\renewcommand{\thefigure}{\Roman{figure}}

\section{Experiments Setting}
\label{sec:supp_expriments}
All experiments for \textbf{IAC} model on the 3 tasks (Photo Retouching, Exposure Correction, and White Balance Editing) were conducted on a single Nvidia A100 GPU. Next, we will provide a detailed explanation of the experimental settings and training details for each task.

\subsection{Photo Retouching Setting}

The MIT-Adobe FiveK~\cite{fivek_dataset} dataset contains 5,000 images, of which 4,500 are used for training and the remaining 500 for evaluation.
The training images are uniformly resized to 400$\times$600 and augmented with random flip and rotation. The training process uses the Adam optimizer with an initial learning rate of $1e^{-5}$ and a weight decay set to 0.0002. The model is trained for a total of 100 iterations, accompanied by a cosine annealing learning strategy.

The loss function between predicted image $F^{-1}(L(F(x)))$ and ground truth image $\hat{x}$  is a mixed loss function $\mathcal{L}_{mix}$ consisting of smooth L1 loss and VGG loss~\cite{Perceptual_loss}:

\begin{equation}
   \mathcal{L}_{mix} = \mathcal{L}_{1smooth} + 0.04 \cdot \mathcal{L}_{vgg}.
\label{eq:total_function}
\end{equation}

\subsection{Exposure Correction Setting}

The exposure correction \textbf{ME} dataset~\cite{Afifi_2020_CVPR_exposure_dataset}  contains 24,330 images,
which divided into 17,675 training images, 750 validation images, and 5905 test images. For the exposure correction task, the training images are cropped into 256$\times$256 patches and augmented  with random  flip and rotation. We also adopt Adam optimizer same as photo retouching task, with an initial learning rate of $2e^{-5}$ and a weight decay set to 0.0001. The model is trained for a total of 20 iterations, also accompanied by a cosine annealing learning strategy. And the loss fucntion we used in exposure correction task is L1 loss function.

\subsection{White Balance Editing Setting}

The White Balance Editing dataset~\cite{afifi2019colour_knn}, Rendered WB, includes two sets: \textbf{Set1} containing 62,535 images and \textbf{Set2} containing 2,881 images. We use 12,000 images from \textbf{Set1} for training. The training settings are the same as for the exposure correction tasks, except the number of training epochs is set to 100. We also adopt the L1 loss function for this task.


\section{Ablation Analyse}

\subsection{Curve Dimension Ablation}

\begin{figure*}[t]
    \centering
    \includegraphics[width=0.95\linewidth]{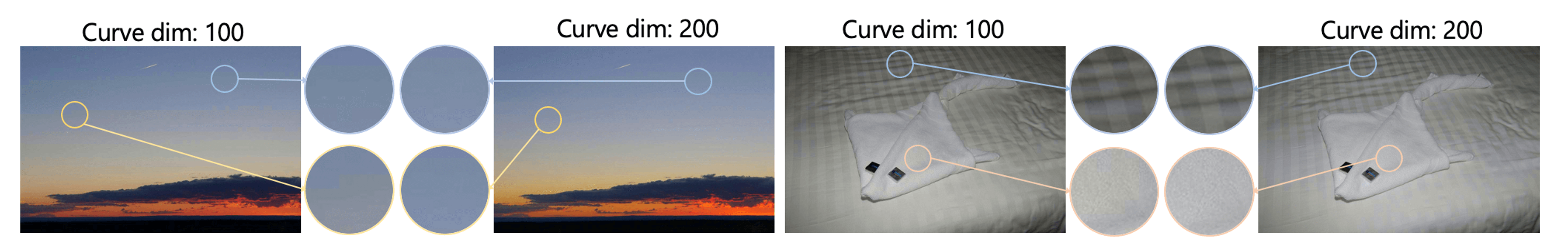}
    \caption{The ablation analyse of the curve dim's effect.}
    \label{fig:dims}
    
\end{figure*}

We conducted an ablation analysis on the dimensionality of curves on the exposure correction \textbf{ME}~\cite{Afifi_2020_CVPR_exposure_dataset}  dataset, investigating the impact of the dimensions of $\left\{\textit{curve}_1, \textit{curve}_2, \textit{curve}_3\right\}$ on the experimental results. The experimental results are shown in Table.~\ref{tab:dim}, from which we can observe that setting the dimension to 200 is a rather reasonable choice. Meanwhile, as shown in Fig.~\ref{fig:dims}, setting a lower dimension easily leads to pixelation in the images.

\subsection{Network Structure Ablation}

In our default experiments, we set the dimension size in ConvNext~\cite{liu2022convnet} block to 32, in the ablation study on photo retouching dataset~\cite{fivek_dataset}, we attempted other dimension numbers such as 16, 24, and 64, as shown in Table.~\ref{tab:size}. From the perspective of parameter count/FLOPs and overall performance, we found that setting it to 24 or 32 is a more reasonable choice. 
Furthermore, in the original network, we attempted to use three parallel branches to learn three sets of coordinates and their corresponding curves. Here, we also tried putting the three sets of coordinates and curves into a single branch for learning. However, we found that this approach led to a significant decrease in performance, as shown in Table.~\ref{tab:size} (``unified branch'').

\begin{table}[t]
\caption{Ablation analyse on the curve dimension.}
\centering
\begin{tabular}{c|ccccc}
\hline
\hline
     & dims 50 & dims 100 & dims 150 & dims 200 & dims 250 \\ \hline
PSNR & 17.67   & 20.45    & 20.88    & 21.22    & \textbf{21.24}    \\ \hline
SSIM & 0.634   & 0.795    & 0.832    & \textbf{0.850}    & 0.849    \\ \hline
\hline
\end{tabular}
\label{tab:dim}
\end{table}

\begin{table}[h]
\caption{Ablation analyse on the ConvNext~\cite{liu2022convnet} block convolution dimension.}
\centering
\begin{tabular}{c|cccc|c}
\hline
\hline
           & size 16 & size 24 & size 32 & size 64 & unified branch (size 32) \\ \hline
PSNR       & 24.01   & 24.81   & 25.02   & 25.01   & 24.12                    \\ \hline
SSIM       & 0.872   & 0.897   & 0.902   & 0.895   & 0.865                    \\ \hline
parameters & 16.7K   & 28.9K   & 39.7K   & 97.8K   & 25.4K                    \\ \hline
Flops      & 0.72 G  & 1.98 G  & 3.25 G  & 7.89 G  & 3.04 G                   \\ \hline
\hline
\end{tabular}
\label{tab:size}
\end{table}

\end{document}